\documentclass[runningheads,a4paper]{llncs}
\usepackage{amssymb}
\usepackage{graphicx}

\usepackage{wrapfig}
\usepackage{url}
\usepackage{soul}
\usepackage[usenames, dvipsnames]{color}
\usepackage{colortbl}
\usepackage{multirow}

\newcommand{\hlc}[2][yellow]{{\sethlcolor{#1} \hl{#2}}}

\newcommand{\keywords}[1]{\par\addvspace\baselineskip
\noindent\keywordname\enspace\ignorespaces#1}

\begin{document}

\mainmatter  

\title{M2D: Monolog to Dialog Generation for Conversational Story Telling}


%
%
\author{Kevin K. Bowden \and Grace I. Lin \and Lena I. Reed \and \\ Jean E. Fox Tree \and Marilyn A. Walker}
%

\institute{Natural Language and Dialog Systems Lab\\
University of California, Santa Cruz\\
\path{{kkbowden, glin5, lireed, foxtree, mawwalker}@ucsc.edu}}


%
%

\maketitle

\begin{abstract}
Storytelling serves many different social functions, e.g. stories are
used to persuade, share troubles, establish shared values, learn
social behaviors, and entertain.  Moreover, stories are often told
conversationally through dialog, and previous work suggests that
information provided dialogically is more engaging than
when provided in monolog.  In this paper, we present algorithms for
converting a deep representation of a story into a dialogic
storytelling, that can vary aspects of the telling, including the
personality of the storytellers. We conduct several experiments to
test whether dialogic storytellings are more engaging, and whether
automatically generated variants in linguistic form that correspond to
personality differences can be recognized in an extended storytelling
dialog.

\keywords{ICIDS, Dialog, Natural Language Generation, Personality, Conversational Storytelling, Analysis and Evaluation of Systems}

\end{abstract}

\section{Introduction}

Storytelling serves many different social functions, e.g. stories are
used to persuade, share troubles, establish shared values, learn
social behaviors, and entertain
\cite{ryokaiVaucelleCassell03,PennebakerSeagal99}.
Moreover, stories are often told conversationally through dialog
\cite{Thorneetal07,Thorne87} where the telling of a story is shaped by the personality
of both the teller and the listener. For example, extraverted friends
actively engage one another in constructing the action of the story by
peppering the storyteller with questions, and by asking the listener
to guess what happened \cite{Thorneetal07,Thorne87}.  
Thus the same story can be told in many different ways, often
achieving different effects
\cite{PasupathiHoyt09,ThorneMcLeanLawrence04}.

A system capable of telling a story and then retelling it in different
settings to different audiences requires two components:
1) a deep representation of the {\sc story} and
2) algorithms that render the story content as different {\sc discourse} instantiations.
A deep representation of the story's content,
often called the {\sc story} or {\sc fabula}, must specifies 
the events, characters, and props of the story, as well as relations among them,
including reactions of characters to story events.
This is accomplished through {\sc est} \cite{Rishesetal13},
a framework that bridges the story annotation tool {\sc scheherazade} and 
a natural language generator (NLG).

The {\sc discourse} representation is the surface rendering of
the {\sc fabula}, an instantiated expressive telling of a story as a
stream of words, gestures or expressions
\cite{bal81,genette83,prince73,propp28}.
This paper presents {\sc m2d}, a framework with algorithms that manipulate the story content to retell 
the story as a conversational dialog between two people.
An example of the original, monologic, and dialogic telling of the Garden Story is shown in Figure \ref{garden-dialog}.
Note that highlighted areas indicate the same information being presented differently at different stages.

\begin{figure}
\centering
\begin{scriptsize}
\vspace{-0.2in}
\begin{tabular}{p{2.3in} | p{2.4in}}
\hline
\multicolumn{2}{c}{\cellcolor[gray]{0.9}{\bf Garden Story: Original}} \\
\hline
\multicolumn{2}{p{4.75in}}{
Today when I arrived at my community garden plot, it actually looked like a garden. Not a weedy mess with maybe some stuff growing in it if you know where to look.
We had hit the typical mid-summer mess of fast-growing weeds and no time to do anything about it. \hlc[YellowOrange]{Plus all the rain had made a huge swamp and it was hard to get a moment to work when it wasn't actively pouring.} I put in a bunch of time this past week, and it's paying off.
Along with free-standing non-weed-choked plants, I have now re-planted three of the beds with salad greens, spinach, and chard. And while the viability of the seeds was questionable, \hlc[YellowOrange]{I accidentally unearthed some from the bed I planted 2 days ago and they had already started to sprout!} {\it ... etc.}
%
} \\
\hline \hline
\cellcolor[gray]{0.9}{\bf Garden Story: Monolog generated by {\sc est}}
& \cellcolor[gray]{0.9}{\bf Garden Story: Dialog generated by {\sc m2d}} \\
\hline
\multirow{7}{2.3in}{
\hlc[YellowOrange]{The radishes charmed the butterflies. The communal garden was weedy. It rained.
The communal garden was swampy.} It rained. The productive gardener planted the
plants. The gardener planted the chards the lettuces and the spinach. \hlc[YellowOrange]{The
pleased gardener did not expect for the chards the lettuces and the spinach to
grow.}
The chards the lettuces and the spinach sprouted. The gardener mistakenly
dug the chards the lettuces and the spinach. The surprised gardener saw for the
chards the lettuces and the spinach to sprout. The communal garden was not
weedy. The communal garden was not swampy. The communal garden was productive.
The gardener was proud. The eager gardener wanted to reap the lettuces. The
radishes were droopy. The gardener planned to remove the radishes. The
thoughtful gardener thought the flowers charmed the butterflies.}
 &
 \hlc[YellowOrange]{{\bf S1:} 
The radishes charmed the butterflies. Technically, the damn communal garden was
not weedless. Err ... it rained. Great right, it was really swampy. } \\
& S2: 
Yeah, it rained. The productive gardener planted the plants. I mean, she planted
the chards the lettuces and the spinach. \\
& \hlc[YellowOrange]{{\bf S1:} 
She did not expect for them to grow. The chards the lettuces and the spinach
sprouted, didn't they?} \\
& {\bf S2:} 
Unfortunately, the gardener mistakenly dug them. She saw for the chards the lettuces
and the spinach to sprout. \\
& {\bf S1:} 
The communal garden was not weedy. It was pretty productive and not swampy. \\
& S2: 
Mmhm ... the gardener was proud and wanted to reap the lettuces. I thought
everybody knew that they were quite droopy? The radishes were droopy. \\
& {\bf S1:} 
I see, well, she planned to remove them. The thoughtful gardener thought the flowers charmed
the butterflies.\\
\hline
\end{tabular}
\end{scriptsize}
\vspace{-0.2in}
\caption{Garden Story: Original Version and Monologue/Dialog Generation. \label{garden-dialog}}
Highlighted areas indicate examples of the same information.
\vspace{-0.2in}
\end{figure}


We build on the publicly available PersonaBank corpus \footnote{Available from {\tt nlds.soe.ucsc.edu/personabank}}, which provides
us with the deep story representation and a lexico-syntactic
representation of its monologic retelling
\cite{LukinWalker15}. 
PersonaBank consists of a corpus
of monologic personal narratives from the ICWSM Spinn3r Corpus
\cite{Burtonetal09} that are annotated with a deep story
representation called a {\sc story intention graph}
\cite{Lukinetal16}.  
After annotation, the stories are run through the {\sc est} system to generate  
corresponding deep linguistic structure representations.
{\sc m2d} then takes these representations as input and creates {\bf dialog with different character voices}.
We identify several stories by hand as good candidates for dialogic tellings
because they describe events or experiences that two people could
have experienced together.  

Our primary hypothesis is {\bf H1}: Dialogic tellings of stories are more engaging than monologic tellings.
We also hypothesize that good dialog requires the use of 
narratological variations such as direct speech, first person, and
focalization \cite{LukinWalker15}. 
Moreover, once utterances are rendered as
first-person with direct speech, then character voice becomes relevant, because
it does not make sense for all the characters and the narrator to talk
in the same voice.  Thus our primary hypothesis {\bf H1} entails two
additional hypotheses {\bf H2} and {\bf H3}: 

\begin{itemize}
\item[] {\bf H2:} Narratological variations such as direct speech, first
person, and focalization will affect a readers engagement with a story. 
\item[] {\bf H3:} Personality-based variation is a key aspect of
  expressive variation in storytelling, both for narrators and story
  characters.  Changes in narrator or character voice may affect
  empathy for particular characters, as well as engagement
and memory for a story.  
\end{itemize}

Our approach to creating different character voices 
is based on the {\bf Big Five} theory of personality
\cite{AllportOdbert36,Goldberg90}. It provides a useful level of abstraction
(e.g., extraverted vs. introverted characters)
that helps to generate language and to guide
the integration of verbal and nonverbal behaviors
\cite{MairesseWalker10,Neffetal10,Huetal15}.  


To the best of our knowledge, our work is the first to develop and
evaluate algorithms for automatically generating different dialogic
tellings of a story from a deep story representation, and the first to
evaluate the utility and effect of parameterizing the style of 
speaker voices (personality) while telling the story.


\section{Background and Motivation}
\label{relwork-sec}

Stories can be told in either dialog or as a monolog, and in many
natural settings storytelling is conversational
\cite{Bavelasetal00}. Hypothesis {\bf H1} posits that dialogic
tellings of stories will be more engaging than monologic
tellings.  In storytelling and at least some educational settings,
dialogs have cognitive advantages over monologs for learning and
memory. Students learn better from a verbally interactive agent than
from reading text, and they also learned better when they interacted
with the agent with a personalized dialog (whether spoken or written)
than a non-personalized monolog \cite{Morenoetal00}.  Our experiments
compare different instances of the dialog, e.g. to test whether
more realistic conversational exchanges affects whether people become
immersed in the story and affected by it.

Previous work supports {\bf H2}, claiming that direct, first-person
speech increases stories' drama and memorability
\cite{Schiffrin81,Tannen89}. 
Even when a story is told as a
monolog or with third person narration, dialog is an essential part
of stortelling: in one study of 7 books, between 40\% and 60\% of
the sentences were dialog \cite{deHaan96}.  
In general narratives are
mentally simulated by readers \cite{Speeretal09}, but readers also
enact a protagonist's speech according to her speech style, reading
more slowly for a slow-speaking protagonist and more quickly for a
fast-speaking protagonist, both out-loud and silently
\cite{YaoScheepers11}.  However, the speech simulation only occurred
for direct quotation (e.g. {\it She said ``Yeah, it rained''}), not
indirect quotation (e.g. {\it She said that it had rained}). 
Only direct quotations activate voice-related parts of
the brain \cite{YaoScheepers11}, as they create a more vivid
experience, because they express enactments of previous events,
whereas indirect quotations describe events \cite{WadeClark93}.

Several previous studies also suggest {\bf H3}, that personality-based
variation is a key aspect of storytelling, both for narrators and
story characters.  Personality traits have been shown to affect how
people tell stories as well as their choices of stories to tell
\cite{McAdamsetal04}. And people also spontaneously encode trait
inferences from everyday life when experiencing narratives, and they
derive trait-based explanations of character's behavior
\cite{RappGerrigPrentice01,Ross77}. Readers use these
trait inferences to make predictions about story outcomes and prefer
outcomes that are congruent with trait-based models
\cite{RappGerrigPrentice01}. The finding that the behavior of the story-teller is affected by the
personality of both the teller and the listener also motivates our
algorithms for monolog to dialog generation
\cite{Thorne87,Thorneetal07}. Content allocation should be controlled by the personality of
the storyteller (e.g. enabling extraverted agents to be more verbose than introverted agents). 

Previous
work on generation for fictional domains has typically combined {\sc
  story} and {\sc discourse}, focusing on the generation of {\sc
  story} events and then using a direct text realization strategy to
report those events
\cite{meehan76}. This approach cannot
support generation of different tellings of a story
\cite{peinadoGervas06}. 
Previous work on generating
textual dialog from monolog suggests the utility of adding extra
interactive elements (dialog interaction) to storytelling and some
strategies for doing so
\cite{Andreetal00,StoyanchevPiwek10,PiwekStoyanchev10}.  
In addition, expository or persuasive content rendered as dialog is
more persuasive and memorable
\cite{VanDeemteretal08,Piweketal07,PiwekStoyanchev11}.
None of this previous work attempts to generate
dialogic storytelling from original monologic content.



\section{M2D: Monolog-to-Dialog Generation}
\label{m2d-sys}

\begin{wrapfigure}{r}{0.5\textwidth}
\centering
\vspace{-.2in}
\includegraphics[width=.49\textwidth]{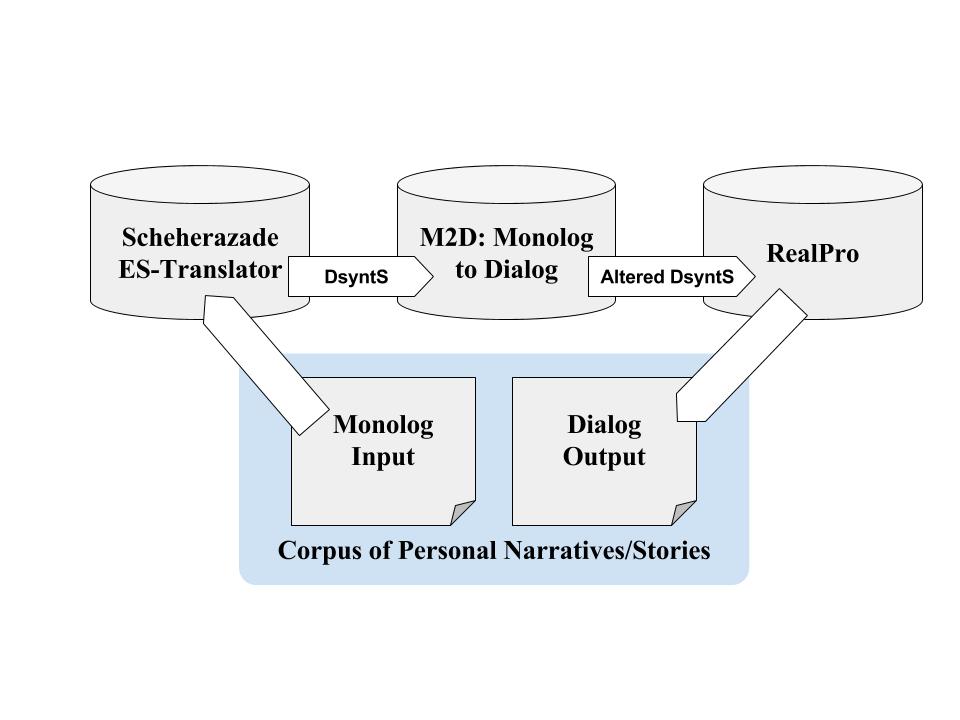}
\caption{{\sc m2d} Pipeline Architecture.}
\label{pipeline-fig}
\vspace{-.3in}
\end{wrapfigure}

Figure~\ref{pipeline-fig} illustrates the architecture of {\sc m2d}. 
The {\sc est} framework produces a story annotated by {\sc scheherazade} as a list of Deep Syntactic Structures ({\bf DsyntS}). DsyntS, the input format for the surface realizer RealPro \cite{LavoieRambow97,Melcuk88}, is a dependency-tree structure where each node contains the lexical information for the important words in a sentence.
Each sentence in the story is represented as a DsyntS.


{\sc m2d} converts a story (as a list of DsyntS) into different versions of a two-speaker dialog
using a parameterizable framework. The input
parameters control, for each speaker, the allocation of content, 
the usage of questions of different forms, and the usage 
of various pragmatic markers (Table \ref{table:exmarkers}).  
We describe the {\sc m2d} parameters in more details below.

\begin{table}[h!]
\caption{Dialog Conversion Parameters \label{table:exmarkers}}
\vspace{-.1in}
\centering
\begin{scriptsize}
\begin{tabular}{l | p{3.6in}}
\hline 
{\bf Parameter} & {\bf Example} \\
\hline
\multicolumn{2}{c}{\bf \cellcolor[gray]{0.9}Aggregation}  \\
\hline 
Merge short sents & The garden was swampy, and not productive.\\ 
Split long sents & The garden was very swampy because it rained. The garden is very large, and has lots of plants. \\ 
\hline 
\multicolumn{2}{c}{\bf \cellcolor[gray]{0.9}Coreference}  \\
\hline 
Pronominalize & The gardener likes to eat  apples from his orchard. They are red.\\  
\hline 
\multicolumn{2}{c}{\bf \cellcolor[gray]{0.9}Pragmatic Markers}  \\
\hline 
Emphasizer\_great & Great, the garden was swampy.  \\
Downer\_kind\_of & The garden was kind of swampy.  \\
Acknowledgment\_yeah & Yeah, the garden was swampy. \\
Repetition & $S_1$: The garden was swampy.  \\
              & $S_2$: Yeah, the garden was swampy. \\
Paraphrase    & $S_2'$: Right, the garden was boggy.  \\
\hline 
\multicolumn{2}{c}{\bf \cellcolor[gray]{0.9}Interactions}  \\
\hline 
Affirm Adjective & $S_1$: The red apples were tasty and $---$  \\
         & $S_2$: Just delicious, really.  \\
         & $S_1$: Yeah, and the gardener ate them.  \\ 
Correct Inaccuracies & $S_2$: The garden was not productive and $---$.\\
        & $S_1$: I don't think that's quite right, actually. I think the garden was productive. \\
\hline 
\multicolumn{2}{c}{\bf \cellcolor[gray]{0.9} Questions}  \\
\hline 
Provoking & I don't really remember this part, can you tell it? \\
With\_Answer & $S_1$: How was the garden?\\
         & $S_2$: The garden was swampy.  \\
Tag & The garden was swampy, wasn't it? \\ 
\hline 
    \end{tabular}
\end{scriptsize}
\vspace{-.3in}
\end{table}

\noindent
{\bf Content allocation:} 
We allocate the content of the original 
story between the two speakers using a content-allocation parameter
that ranges from 0 to 1. 
A value of .5 means that the content is equally split between 2 speakers.
This is motivated by the fact that, for example, extraverted speakers
typically provide more content than intraverted speakers
\cite{MairesseWalker10,Thorne87}.

\noindent{\bf  Character and Property Database:} We use
the original source material for the story to infer information about actors,
items, groups, and other properties of the story, using the
information specified in the DsyntS.
We create actor objects for each character
and track changes in the actor states as the story proceeds,
as well as changes in basic properties such as their 
their body parts and possessions.

\noindent{\bf Aggregation and Deaggregation:} We break apart long
DsyntS into smaller DsyntS, and then check where we can merge small
and/or repetitious DsyntS. We believe that deaggregation will
improve our dialogs overall clarity while aggregation will make our
content feel more connected \cite{Lukinetal15}.

\begin{wrapfigure}{R}{0.5\textwidth}
\centering
\vspace{-.2in}
\includegraphics[width=0.49\textwidth]{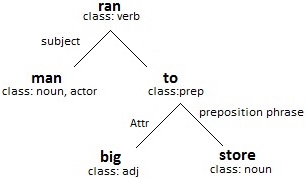}
\vspace{-.1in}
\caption{The DsyntS tree for \textit{The man ran to the big store.} \label{fig:question_tree}}
\vspace{-.3in}
\end{wrapfigure}

\noindent{\bf Content Elaboration:} In natural dialog, speakers often repeat
or partially paraphrase each other, repeating the same content in multiple ways.
This can be a key part of entrainment. Speakers may 
also ask each other questions thereby setting up frames for interaction \cite{Thorne87}. 
In our framework, this involves duplicating content in a single DsyntS by either 1) generating
a question/answer pair from it and allocating the content across speakers, or 2) duplicating
it and then generating paraphrases or repetitions across speakers. 
Questions are generated by performing a series of pruning operations based on the class of the selected node and the relationship with its parent and siblings. For example, if {\sc store} in Figure \ref{fig:question_tree} is selected, we identify this node as our question. The class of a node indicates the rules our system must follow when making deletions. Since {\sc store} is a noun we prune away all of the {\sc attr} siblings that modify it. By noticing that it is part of a prepositional phrase, we are able to delete {\sc store} and use {\sc to} as our question, generating \textit{The man ran where?}.

\noindent{\bf Content Extrapolation:} We make use of the deep
underlying story representation and the actor database to make
inferences explicit that are not actually part of the original
discourse. For example, the actor database tracks aspects of a
character's state. By using known antonyms of the adjective defining
the current state, we can insert content for state changes, i.e.  the
alteration from {\it the fox is happy} to {\it now, the fox is sad},
where the fox is the actor and happiness is one of his
states. This also allows us to introduce new dialogic interactions by
having one speaker ask the other about the state of an actor, or make
one speaker say something incorrect which allows the second speaker to
contradict them: {\it The fox was hungry} followed by {\it No, he
  wasn't hungry, he was just greedy.}

\noindent{\bf Pragmatic Markers:} We can also insert pragmatic markers
and tag questions as described in Table~\ref{table:exmarkers}.
Particular syntactic constraints are specified for each pragmatic
marker that controls whether the marker can be inserted at all \cite{MairesseWalker10}.  
The frequency and type of
insertions are controlled by values in the input parameter file. Some
parameters are grouped by default into sets that allow them to be used
interchangeably, such as downtoners or emphasizers. 
To provide us more control over the variability of generated variants, 
specific markers which are by default unrelated can be packaged together and 
share a distributed frequency limit. 
Due to their simplistic nature and low number of constraints, 
pragmatic markers prove to be a reliable source of variation in the systems output.

\noindent{\bf  Lexical Choice:} We can also replace 
a word with one of its synonyms.  This can be driven simply
by a desire for variability, or by lexical choice parameters
such as word frequency or word length.

\noindent{\bf  Morphosyntactic Postprocessing:} The final
postprocessing phase forms contractions and  possessives
and corrects known grammatical errors.

The results of the {\sc m2d} processor are then given as input to RealPro
\cite{LavoieRambow97}, an off-the-shelf surface text realizer. RealPro
is responsible for enforcing English grammar rules, morphology,
correct punctuation, and inserting functional words in order to
produce natural and grammatical utterances.

\section{Evaluation Experiments}
\label{eval-sec}

We assume H1 on the basis of previous experimental work, and test H2
and H3.  Our experiments aim to: (1) establish whether and to what
degree the {\sc m2d} engine produces natural dialogs; (2) determine how the
use of different parameters affect the user's engagement with the
story and the user's perceptions of the naturalness of the dialog; and
(3) test whether users perceive personality differences that are
generated using personality models inspired by previous work.  All
experimental participants are pre-qualified Amazon Mechanical Turkers 
to guarantee that they provide detailed and thoughtful comments.

We test users' perceptions of naturalness and engagement
using two stories: the Garden story (Figure~\ref{garden-dialog}) and 
the Squirrel story (Figure~\ref{squirrel-dialog}).
For each story, we generate {\bf three} different dialogic
versions with varying features: 
\begin{description}
\item[{\sc m2d-est}] renders the output from {\sc est} as a dialog by
allocating the content equally to the two speakers. {\bf No variations} of sentences are introduced.
\item[{\sc m2d-basic}] consists of transformations required to produce a minimally natural
dialog. First we apply {\bf pronominalization}  to
replace nouns with their pronominal forms when telling the story in
the third person.  We then manipulate {\bf sentence length} by
breaking very long sentences into shorter ones, or by combining
repetitious short sentences into one sentence. 
This is motivated by the fact that utterances in dialog tend to be less formal and use less complex syntactic
structures \cite{Biber91}. 
The last transformation is 
{\bf morphosyntactic postprocessing} as described in Section \ref{m2d-sys}. 
\item[{\sc m2d-chatty}] adds interactive features to {\sc m2d-basic} such as the
insertion of {\bf pragmatic markers} (acknowledgements, disfluencies, hedges) and question-answer generation (Table
\ref{table:exmarkers}).
\end{description}

\begin{figure}[h!]
\centering
\begin{scriptsize}  
\vspace{-0.3in}
\begin{tabular}{|p{4.7in}|}
\hline 
\cellcolor[gray]{0.9}{\bf Squirrel Story: Monolog generated by {\sc est}} \\
\hline
I placed the steely bowl on the deck in order for Benjamin to drink the bowl's
water. The steely bowl was popular. The birds drank the bowl's water. The birds
bathed themselves in the steely bowl. The birds organized themselves on the
deck's railing in order for the birds to wait. The squirrels drank the bowl's
water. The squirrel approached the steely bowl. 
The crazy squirrel was startled
because the squirrel saw the squirrel's reflection. The crazy squirrel leaped
because the squirrel was startled. The crazy squirrel fell over the deck's
railing because the squirrel leaped because the squirrel was startled. The crazy
squirrel held the deck's railing with the squirrel's paw. The squirrel's paw
slipped off the deck's railing. The crazy squirrel fell. 
\\
\hline

\cellcolor[gray]{0.9}{\bf Squirrel Story: Dialog generated by {\sc m2d}} \\
\hline
{\bf S1:}  I placed the steely bowl on the deck. Benjamin drinks its water.\\
{\bf S2:}  The steely bowl was very popular, technically. Typical. The birds drank its
water. They bathed themselves in the steely bowl! The birds organized themselves
on the deck's railing in order for them to wait.\\
{\bf S1:} 
 Yeah, the birds organized themselves on the deck's railing in order for them to wait.
The squirrels drank it's water. The squirrel approached the steely bowl.\\
{\bf S2:} 
 Technically, it was startled. I am delighted to say that the squirrel saw its
reflection. It jumped, actually and was startled. Basically, the squirrel was
literally startled and fell over the deck's railing. It leaped because it was
startled.\\
{\bf S1:} 
 The crazy squirrel held the deck's railing with its paw.\\
{\bf S2:} 
 Its paw slipped off its railing pal. The particularly crazy squirrel fell
mate.\\
\hline
\end{tabular}
\end{scriptsize}
\vspace{-0.2in}
\caption{Squirrel Story: Monolog/Dialog Generation. \label{squirrel-dialog}}
\vspace{-0.2in}
\end{figure}

Each pairwise
comparison is a Human Intelligence Task (HIT; a question that needs an answer),
yielding 6 different HITs. We used 5 annotators (Turkers) per HIT to 
rate the levels of engagement/naturalness on a
scale of 1-5, followed by detailed comments justifying their ratings.

We create several subsets of features that work well together
and recursively apply random feature insertion 
to create many different output generations. 
These subsets
include the types of questions that can be asked, different speaker
interactions, content polarity with repetition options, pragmatic
markers, and lexical choice options. A restriction is
imposed on each of the subgroups, indicating the maximum number of
parameters that can be enabled from the associated
subgroup. This results in different styles of speaker
depending  on which subset of features is chosen. A speaker who
has a high number of questions along with hedge pragmatic markers will
seem more inquisitive, while a speaker who just repeats what the other
speaker says may appear to have less
credibility than the other speaker. We plan to explore
particular feature groupings in future work to identify specific
dialogic features that create a strong perception of personality.

\subsection{M2D-EST vs. -Basic vs. -Chatty}

\begin{wrapfigure}{R}{3in}
\vspace{-.3in}
\begin{minipage}{.3\textwidth}
  \centering
  \includegraphics[width=.9\linewidth]{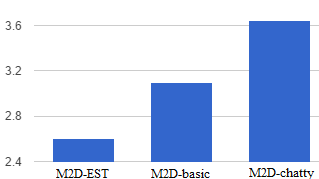}
  \caption{Mean scores for engagement.}
  \label{fig:mean_engage}
\end{minipage}%
~~
\begin{minipage}{.3\textwidth}
  \centering
  \includegraphics[width=.9\linewidth]{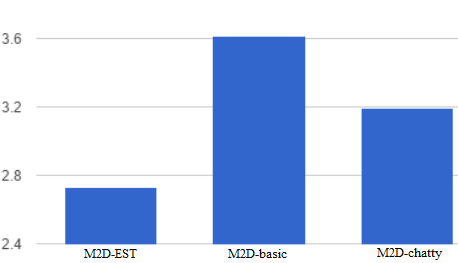}
  \caption{Mean scores for naturalness.}
  \label{fig:mean_natural}
\end{minipage}
\vspace{-.3in}
\end{wrapfigure}

The perceptions of engagement given different versions of the dialogic story
is shown in Figure \ref{fig:mean_engage}.
A paired t-test comparing {\sc m2d-chatty} to {\sc m2d-est} shows that
increasing the number of appropriate features makes the dialog more
engaging (p $=$ .04, df$=$ 9).  However there are no statistically
significant differences between {\sc m2d-basic} and {\sc m2d-est},
or between {\sc m2d-basic} and {\sc m2d-chatty}.
Comments by Turkers suggest that the {\sc m2d-chatty} speakers 
have more personality because they use
many different pragmatic markers, such as questions and other dialogically
oriented features.  

The perception of naturalness across the same set of dialogic stories is shown
in Figure \ref{fig:mean_natural}.
It shows that {\sc m2d-basic} was rated higher than {\sc m2d-est}, and
their paired t-test 
shows that {\sc m2d-basic} (inclusion of pronouns and agg- and deaggregation)
has a positive impact on the naturalness of a dialog (p $=$ .0016, df $=$ 8).
On the other hand,  
{\sc m2d-basic} is preferred over {\sc m2d-chatty}, 
where the use of pragmatic markers in {\sc m2d-chatty} was often noted as unnatural.

%

\subsection{Personality Models}

A second experiment creates a version of {\sc m2d} called {\sc m2d-personality}
which tests whether users perceive the personality that
{\sc m2d-personality} intends to manifest. We use 4 different stories
 from the PersonaBank corpus \cite{Lukinetal16} and create
introverted and extroverted personality models, partly by drawing on
features from previous work on generating personality
\cite{MairesseWalker10}. 

\begin{wraptable}{R}{3in}

\vspace{-.1in}
\caption{Feature Frequency for Extra. vs. Intro.}
\label{table:personality_models}
\centering
\begin{scriptsize}
Not all lexical instantiations of a feature are listed.\\
\begin{tabular}{ l | l | l }
\hline 
       \textbf{Parameter} & \textbf{Extra.} & \textbf{Intro.}\\
\hline 
\multicolumn{3}{c}{\bf \cellcolor[gray]{0.9} Content allocation}  \\
                  Content density & high & low \\
\hline 
\multicolumn{3}{c}{\bf \cellcolor[gray]{0.9} Pragmatic markers}  \\
Adjective softeners & low & high \\
Exclamation & high & low \\
Tag Questions & high & low \\
Acknowledgments: {\it Yeah, oh God} & high & low\\ 
Acknowledgments: {\it I see, well, right} & low & high\\               
Downtoners: {\it Sort of, rather, quite, pretty} & low & high \\
Downtoners: {\it Like} & high & low \\
Uncertainty: {\it I guess, I think, I suppose} & low & high \\        
Filled pauses: {\it Err..., Mmhm...} & low & high \\
Emphasizers: {\it Really, basically, technically} & high & low \\
In-group Markers: {\it Buddy, pal} & high & low \\
\hline 
\multicolumn{3}{c}{\bf \cellcolor[gray]{0.9} Content elaboration} \\
Questions: Ask \& let other spkr answer & high & low\\
Questions: Rhetorical, request confirmation & low & high\\                   
Paraphrase & high & low\\
Repetition & low & high\\ 
Interactions: Affirm adjective & high & low\\
Interactions: Corrections & high & low\\  
\hline 
\multicolumn{3}{c}{\bf \cellcolor[gray]{0.9} Lexical choice} \\
Vocabulary size & high & low\\ 
Word length & high & low \\
\hline 
\end{tabular}
\end{scriptsize}
\hspace{2in}

\vspace{-.3in}
\end{wraptable}

We use a number of new dialogic features in our personality
models that increase the level of interactivity and entrainment, such
as asking the other speaker questions or entraining on their
vocabulary by repeating things that they have said. Content allocation
is also controlled by the personality of the speaker, so that
extraverted agents get to tell more of the content than introverted
agents. 

We generate two different versions of each dialog, an extroverted and an introverted speaker
(Table~\ref{table:personality_models}).
Each dialog also has one speaker who uses a default
personality model, neither strongly introverted or extraverted.  This
allows us to test whether the perception of the default personality
model changes depending on the personality of the other speaker. We
again created HITs for Mechanical Turk for each variation.  The
Turkers are asked to indicate which personality best describes the
speaker from among extroverted, introverted, or none, and then explain
their choices with detailed comments.  The results are shown in
Table~\ref{table:general_TT}, where Turkers correctly identified
the personality that {\sc m2d-personality} aimed to manifest 88\% of the time.

\begin{table}
\centering
\begin{minipage}{0.5\textwidth}
	\centering
	\caption{Personality Judgments \label{table:general_TT}}
	\vspace{-.1in}
	\begin{tabular}{lccc}
	\hline
	& Extro & Intro & None\\
	\hline
	Extro & 16 & 0 & 0\\
	Intro & 3 & 12 & 1\\ 
	\hline
	\end{tabular}
\end{minipage}%
\begin{minipage}{0.5\textwidth}
	\centering
	\caption{Default Personality Judgments \label{table:none_TT}}
	\vspace{-.1in}
    	\begin{tabular}{lccc}
    	\hline 
    	& Extro & Intro & None  \\
    	\hline 
    	Other is Extro & 1 & 7 & 8\\
    	Other is Intro & 8 & 1 & 7\\ 
    	\hline 
    	\end{tabular}	
\end{minipage}
\vspace{-.2in}
\end{table}

Turkers' comments noted the differential use of pragmatic markers, 
content allocation, asking questions, and
vocabulary and punctuation.  
The {\bf extroverted} character
was viewed as more dominant, engaging, excited, and
confident. These traits were tied to the
features used: exclamation marks, questions asked, exchanges between
speakers, and pragmatic markers
(e.g., {\it basically}, {\it actually}).

The {\bf introverted} character was generally
timid, hesitant, and keeps their thoughts to
themselves. Turkers noticed that the
introverted speaker was allocated less content, the tendency to
repeat what has already been said, and the use of different
pragmatic markers 
(e.g. {\it kind of}, {\it I guess}, 
{\it Mhmm}, {\it Err...}).

Table \ref{table:none_TT} shows Turker judgements for the speaker in
each dialog who had a default personality model. 
In 53\% of the trials, our participants picked a personality other than ``none" for
the speaker that had the default personality. Moreover, in 88\% of these
incorrect assignments, the personality assigned to the speaker was the
opposite of the personality model assigned to the other speaker. These
results imply that when multiple speakers are in a conversation,
judgements of personality are {\bf relative} to the other speaker. For example, 
an introvert seems more introverted in the presence of an extravert,
or a default personality may seem introverted in the presence of an extravert.

\section{Discussion and Future Work}
\label{results-sec}

We hypothesize that dialogic storytelling may produce more engagement
in the listener, and that the capability to render a story as dialog
will have many practical applications (e.g. with gestures \cite{Hu16}. 
We also hypothesize that
expressing personality in storytelling will be useful and show how it
is possible to do this in the experiments presented here.  We
described an initial system that can translate a monologic deep syntactic
structure into many different dialogic renderings.

We evaluated different versions of our {\sc m2d} system. The results
indicate that the perceived levels of engagement for a dialogic
storytelling increase proportionally with the density of
interactive features. Turkers commented that the use of pragmatic
markers, proper pronominalization, questions, and other interactions
between speakers added personality to the dialog, making it more
engaging.
In a second experiment, we directly test whether Turkers perceive that
different speaker's personalities in dialog.  We compared introvert, extrovert, and a speaker with a
default personality model. The results show that in 88\% of cases
the reader correctly identified the personality model
assigned to the speaker. The results show 
that the content density assigned to each speaker as well as
the choice of pragmatic markers are strong indicators of the
personality.  Pragmatic markers that most emphasize speech, or
attempt to engage the other speaker are associated with extroverts,
while softeners and disfluencies are associated with
introverts. Other interactions such as correcting false statements and
asking questions also contribute to the perception
of the extroverted personality.

In addition, the perceived personality of the default personality
speaker was affected by the personality of the other speaker.  
The default personality speaker was classified as having a
personality 53\% of the time. In 88\% of
these misclassifications, the personality assigned to the speaker was
the opposite of the other speaker, suggesting that personality
perception is relative in context.

While this experiment focused only on extrovert and introvert, 
our framework contains other Big-Five personality models
that can be explored in the future.
We plan to investigate:
1) the effect of varying feature density on the perception of a personality model,
2) how personality perception is relative in context, and
3) the interaction of particular types of content or dialog acts with perceptions of 
a storyteller's character or personality.
The pragmatic markers are seen as unnatural in some cases. 
We note that our system currently inserts them probabilistically 
but do not make intelligent decisions about using them in 
pragmatically appropriate situations. 
We plan to add this capability in the future.
In addition we will explore new parameters that improves
the naturalness and flow of the story.

\section*{Acknowledgments}
We would like to thank Johnnie Chang and Diego Pedro for their roles as collaborators in the early inception of our system.

%
%


\bibliography{../nl}
\bibliographystyle{plain}

\end{document}